\pdfoutput=1

\documentclass[11pt]{article}

\usepackage{EMNLP2023}

\usepackage{times}
\usepackage{latexsym}

\usepackage[T1]{fontenc}

\usepackage[utf8]{inputenc}

\usepackage{microtype}

\usepackage{inconsolata}

\usepackage{graphicx}
\usepackage{subfig}
\usepackage{appendix}
\usepackage{amsmath,amssymb}
\usepackage{lipsum}

\usepackage{enumitem}

\usepackage{siunitx}

\usepackage{booktabs}
\usepackage{hyperref}

\begin{document}

\title{PRewrite: Prompt Rewriting with Reinforcement Learning}

\author{
Weize Kong$^1$~~~~~~Spurthi Amba Hombaiah$^1$~~~~~~Mingyang Zhang$^1$\\
\textbf{Qiaozhu Mei$^2$~~~~~~Michael Bendersky$^1$}
\\ \\
$^1$Google DeepMind \quad $^2$University of Michigan\\
$^1${\small \texttt{\{weize,spurthiah,mingyang,bemike\}@google.com}} \quad
$^2${\small \texttt{qmei@umich.edu}}
}


\maketitle

\begin{abstract}
Prompt engineering is critical for the development of LLM-based applications. However, it is usually done manually in a ``trial and error'' fashion that can be time consuming, ineffective, and sub-optimal. Even for the prompts which seemingly work well, there is always a lingering question: can the prompts be made better with further modifications?

To address these problems, we investigate automated prompt engineering in this paper. Specifically, we propose PRewrite, an automated method to rewrite an under-optimized prompt to a more effective prompt. We instantiate the prompt rewriter using an LLM. The rewriter LLM is trained using reinforcement learning to optimize the performance on a given downstream task. We conduct experiments on diverse benchmark datasets, which demonstrates the effectiveness of PRewrite.

\end{abstract}
\section{Introduction}
\label{sec:introduction}


With the right prompts, large language models (LLMs) can show impressive performance on various tasks in zero-shot or few-shot settings~\cite{brown2020language,srivastava2022beyond}. 
However, manual prompt engineering is done on a trial-and-error ad-hoc basis and there are limited guiding principles on writing good prompts. 

To address the problems, we investigate methods to automate the process of prompt engineering, often called ``automated prompt engineering'' or ``prompt optimization''. Automated prompt engineering is important due to the wide and fast adoption of LLM applications. Moreover, LLMs themselves are evolving, and as a result,  we also need effective automated methods to update existing prompts to adapt to new models.


Several previous works have explored automated prompt engineering. AutoPrompt~\cite{Shin2020} uses a gradient-based search method to iteratively edit prompts, but requires gradient access to the language model. RLPrompt~\cite{Deng2022} optimizes prompts using reinforcement learning (RL), but often produces uninterpretable gibberish prompts. Also using RL, TEMPERA~\cite{Zhang2022} allows editing prompts based on task input, but its small action space might hinder exploration.
Another common limitation is that they are based on relatively small-size language models like BERT~\cite{Devlin2019} and RoBERTa~\cite{Liu2019}. It is not clear how well the proposed methods can generalize to larger models, especially with API-only model access.

More recent works like APE~\cite{Zhou2023}, OPRO~\cite{Yang2023} and Promptbreeder~\cite{Fernando2023} use larger models from the PaLM 2~\cite{Anil2023} and the GPT\footnote{\url{https://platform.openai.com/docs/models}} model families.
These works leverage LLMs themselves to propose prompt candidates, and search for a better prompt from them via validating performance on a given training dataset. We follow a similar idea but aim to use RL instead of search to improve the optimization process.

\begin{figure}
    \centering
    \includegraphics[width=\columnwidth]{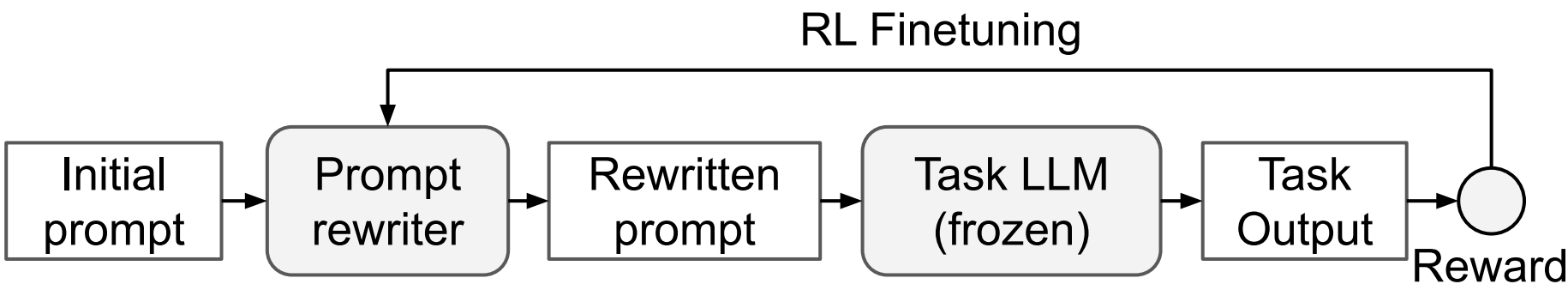}
    \caption{Overview of PRewrite.}
    \label{fig:prewrite_arch}
\end{figure}

In this work, we propose PRewrite, prompt rewriting with reinforcement learning, to address the limitations above. Our idea is to train a prompt rewriter to rewrite an initial under-optimized prompt to a more effective prompt. The prompt rewriter itself is a LLM, trained using RL to optimize for a downstream task. We give an overview in Figure~\ref{fig:prewrite_arch}. 
Specifically, given an initial prompt, the prompt rewriter LLM is instructed to generate a rewritten prompt, which in turn is used by the task LLM to generate the final output. Using a reward computed on the final output against the ground-truth output, the rewriter LLM is finetuned with RL. As compared to previous RL-based methods, PRewrite produces interpretable prompts (cf. RLPrompt), allows unconstrained exploration without manually defined action space (cf. TEMPERA), and leverages larger models (PaLM 2).

Our contributions are summarized as follows:
\vspace*{-8pt}
\begin{itemize}[noitemsep,leftmargin=*]
\item We propose PRewrite, a novel automated prompt engineering approach. It optimizes prompt via rewriting, in an end-to-end manner using reinforcement learning.
\item We develop two rewriting strategies, including one that searches for an optimal rewritten prompt from candidates generated by the RL-trained prompt rewriter (Section~\ref{sec:search}). This often further improves the prompt optimization performance.
\item We conduct experiments on diverse benchmark datasets, which testify the effectiveness of PRewrite and demonstrate its state-of-the-art performance. 
\end{itemize}

\section{PRewrite}
\label{sec:methodology}



\subsection{Problem Formulation}
\label{sec:problem_formulation}
We formulate our prompt rewriting problem more formally in this section. 
Given a text generation task, we denote the task input and output by $x$ and $y$ respectively. To solve the task, one can use a LLM for prediction, $y=\mathtt{LLM}(p)$, where $p$ is the input to the LLM, also known as \textbf{prompt}. Prompts are usually constructed using a template that incorporates the input $x$ and a task \textbf{instruction} $t$. 
For the example, $t$ can be ``\textit{Write a brief answer for the following question}'' for a question answering task. A prompt can be constructed using template \texttt{$p$="\{$t$\}: \{$x$\}"} (Python f-string), where $x$ is the input question. Please refer to Table~\ref{tab:def} in Appendix for a complete example.


Prompt rewriting aims to rewrite a given initial prompt to another prompt $p\textsuperscript{†}=\delta(p)$, in order to optimize the task output. We call the \textbf{rewritten prompt} $p\textsuperscript{†}$. Since prompts can be constructed using an instruction, we simplify the prompt rewriting problem to only rewriting the instruction $t\textsuperscript{†}=\delta(t)$. In fact, most prior works~\cite{Fernando2023, Zhou2023, Yang2023} optimize prompts via optimizing instructions, and do not differentiate between \emph{prompt} and \emph{instruction}. 
In this paper, we use the two terms interchangeably wherever it is clear.

Prompt rewriting can be performed independent or dependent of the task input, i.e., $\delta(\cdot)$ and $\delta(\cdot|x)$. 
We focus on input-independent prompt rewriting, following most prior works. In this case, the instruction is rewritten offline and prompts can be constructed cheaply online using the rewritten instruction.





\subsection{Overview}
\label{sec:overview}

Directly searching for an optimal rewritten prompt is challenging due to the large search space of natural language. So, we propose PRewrite to optimize prompt rewriting, as illustrated in Figure~\ref{fig:prewrite_arch}.

First, the prompt rewriter takes in an initial prompt $p$ and rewrites it to another prompt $p\textsuperscript{†}$. The initial prompt is usually crafted manually and can be sub-optimal. Observing the remarkable capability of LLMs, we instruct a LLM (e.g., PaLM 2-S) with a meta prompt $m$ for rewriting as follows:
\begin{equation}
    p\textsuperscript{†}=\mathtt{LLM}_{R}(\texttt{"\{$m$\}\text{\small \textbackslash nInstruction: }\{$p$\}"}).
    \label{eq:prompt_rewrite}
\end{equation}
We call $\mathtt{LLM}_{R}$, \textbf{rewriter LLM}, which is to be differentiated from the task LLM, used for the end task. We list our meta prompts in Appendix~\ref{app:meta_prompt}.

Second, the rewritten prompt $p\textsuperscript{†}$ is then used by the task LLM to generate the task output. The task LLM is assumed to be a blackbox accessed via API and can be larger than the rewriter LLM.

Third, we compute rewards based on the task output in comparison with the ground-truth output and use reinforcement learning (RL) to finetune the rewriter LLM on a training set (Section~\ref{sec:rl}). This is critical because our meta prompt is very generic. As a result, the rewriter LLM and the rewritten prompt are unlikely to perform well on the downstream task initially.

Lastly, we use the RL-trained prompt rewriter to rewrite the initial prompt according to Equation~\ref{eq:prompt_rewrite} based on two strategies outlined in Section~\ref{sec:search}.








\subsection{Finetuning Rewriter LLM with RL}
\label{sec:rl}
This section provides more details on RL finetuning for our rewriter LLM, which is very similar to other RL-based LLM alignment work~\cite{ouyang2022training}. \textbf{Action space} consists of all tokens in the rewriter LLM's vocabulary, allowing arbitrary text rewriting. \textbf{State} is defined as the concatenation of all the decoded tokens so far. \textbf{Reward} is the task LLM's performance on the downstream task when using the rewritten prompt. We measure this using the end task metric, but also explore other rewards like perplexity and F1 in our experiments (see Appendix~\ref{app:results_rewards}).
We use Proximal Policy Optimization (PPO) \cite{Schulman2017} with KL penalty as the \textbf{RL algorithm} for its robustness. 


A key difference between our work and previous RL-based methods is that we use a capable LLM (PaLM 2-S) as our rewriter model. Because of this, our model is less likely to produce uninterpretable gibberish prompt as in RLPrompt (this can be also attributed to the KL penalty in PPO). We also don't need to define a constrained action space manually as in TEMPERA. 





\subsection{Rewriting via Inference and Search}
\label{sec:search}
Once the rewriter LLM is trained, we use it for prompt rewriting following Equation~\ref{eq:prompt_rewrite}. We design two specific rewriting strategies. For the \textbf{inference} strategy, denoted as \textbf{PRewrite-I}, we set temperature to zero, in which case the model greedily decodes and generates one single rewritten prompt. For the \textbf{search} strategy, denoted as \textbf{PRewrite-S}, we prompt the rewriter LLM $K$-times with temperature=1 to generate a set of prompts, $\{p\textsuperscript{†}_i\}_{i=1}^{K}$. We then select the best $p\textsuperscript{†}_i$ based on their end task performance on a dev dataset.
\section{Experiments \& Analysis}
\label{sec:experiments}


\subsection{Experimental Setup}
\label{sec:experimental_setup}
We evaluate PRewrite on diverse benchmark datasets, spanning from classification with AG News~\cite{Zhang2015} and SST-2~\cite{Wang2018}, question answering with Natural Questions (NQ)~\cite{Kwiatkowski2019} to arithmetic reasoning with GSM8K~\cite{cobbe2021}. 
We use the standard train/dev/test splits. As GSM8K doesn't come with a dev split, so we randomly sample 10\% examples from the train split as the dev split. Data statistics are reported in Appendix~\ref{app:dataset_splits}.

Our initial prompts, prompt templates and meta prompts are listed in Appendix~\ref{app:initial}, \ref{app:prompt_templates} and \ref{app:meta_prompt} respectively.
We experiment with PaLM 2-S and PaLM 2-L~\cite{Anil2023} as the frozen task LLMs with zero temperature. We use PaLM 2-S as the rewriter LLM and set temperature to 1 for both the policy and value model during RL training. We use standard PPO algorithm for online policy optimization with GAE. The model is trained until convergence on the dev set. 
We test both the inference and search strategy for rewriting, denoted as PRewrite-I and PRewrite-S respectively. For PRewrite-S, we search from K=10 rewritten prompts (Section~\ref{sec:search}). 

For baselines, we cite evaluation results for  AutoPrompt~\cite{Shin2020}, RLPrompt~\cite{Deng2022}, and  TEMPERA~\cite{Zhang2022}, out of which the last two are RL-based methods; APE~\cite{Zhou2023}, OPRO~\cite{Yang2023} and Promptbreeder (PB)~\cite{Fernando2023}, which use LLMs of same size as ours.
We report standard metrics on test: accuracy for AG News, SST-2, GSM8K; and Exact Match (EM) for NQ.

\subsection{Results}
\label{sec:results_s}
We first present PRewrite results based on PaLM 2-S task model in Table~\ref{tab:results_s}.
\begin{table}[!ht]
    \centering
    \small
    \begin{tabular}{llllll}
        \toprule
           &  AG News  &  SST-2  &  NQ  &  GSM8K \\
        \midrule
        AutoPrompt  &  65.7  &  75.0  &  -  &  - \\
        RLPrompt  &  77.2  &  90.1  &  -  &  - \\
        TEMPERA  &  81.3  &  92.0  &  -  &  - \\
        \midrule
        Initial prompt  &  76.9  &  96.3  &  24.1  &  29.9 \\
        PRewrite-I  &  84.5  &  96.5  &  29.3  &  52.0 \\
        PRewrite-S  &  85.2  &  96.6  &  30.2  &  53.6 \\
        \bottomrule
    \end{tabular}
    \caption{PRewrite experiment results based on PaLM 2-S task model. The baseline results (top section) are based on RoBERTa-Large task model, cited from TEMPERA~\cite{Zhang2022}. For TEMPERA, non-test-time-editing (No TTE) results are reported.}
    \label{tab:results_s}
\end{table}

\textbf{First}, PRewrite consistently improves over the initial prompts, demonstrating the effectiveness of the proposed method. We repeated the PRewrite experiments 5 times and the results were consistent. We list the rewritten prompts in Table~\ref{tab:rewritten_prompts_s} in Appendix. 
\textbf{Second}, we observe larger improvement for PRewrite when there is more headroom.
For example, the performance gain on SST-2 is minimum, but we observe 80\%, 22\% and 10\% relative improvement with PRewrite on GSM8K, NQ and AG News respectively. 
\textbf{Third}, PRewrite-S consistently shows improvement over PRewrite-I, suggesting that search strategy can be more helpful. We find the two strategies often produce prompts with small differences. For example, ``sentiment classification'' from PRewrite-I and ``sentiment classification \emph{from text}'' for Prewrite-S on SST-2.
\textbf{Lastly}, baseline models underperform PRewrite. However, this can be largely due to the smaller task model, RoBERTa-Large~\cite{Liu2019}, being used for the baselines. That said, it is not straightforward to apply some of the baseline methods on larger models like PaLM 2, especially in case of API-only access.

\begin{table}[!ht]
    \centering
    \small
    \begin{tabular}{ccc|cc}
        \toprule
        APE & OPRO & PB & Initial prompt & PRewrite-S \\
        \midrule
        77.9 & 80.2 & 83.9 & 37.0 & 83.8 \\
        \bottomrule
    \end{tabular}
    \caption{GSM8K experiment results based on PaLM 2-L task model. Baseline results (left section) are cited from PromptBreeder (PB)~\cite{Fernando2023}, also based on PaLM 2-L task model.}
    \label{tab:results_l}
\end{table}

\textbf{Next}, we compare PRewrite with baselines on GSM8K, all based on PaLM 2-L task model in Table~\ref{tab:results_l} (PRewrite-S only due to space constraints). PRewrite-S not only dramatically improves the initial prompt, but also outperforms strong baselines like APE and OPRO, and is on par with Promptbreeder. This result is especially impressive in that the PRewrite setup is relatively simple with minimal customization, in comparison with the baselines. For example, APE proposes to use task input and output to induce instructions for most tasks but has a special treatment to GSM8K. It collects a customized dataset with questions and reasoning steps via prompting InstructGPT for instruction induction -- this is more likely to induce chain-of-thought instructions. Promptbreeder uses 56 mutation prompts and 39 thinking style prompts including ones that contain phrases like \textit{steps required}, \textit{taking a break} or \textit{suggesting explanation}. In comparison, we only use one generic meta prompt for GSM8K (Appendix~\ref{app:meta_prompt}).

We also experiment with different rewards for PRewrite. Please refer to Appendix~\ref{app:results_rewards} for the results.



\subsection{Case Studies}
\label{sec:case_studies}
To showcase the capability of PRewrite, we present rewriting performed by it for two datasets in Table~\ref{tab:case} (see Appendix~\ref{app:initial}, \ref{app:rewritten} for more results). 

For NQ, PRewrite not only learns the task needs a \emph{short} answer, but also impressively adds an in-context example. For GSM8K, PRewrite rewrites the simple initial prompt to a creative chain-of-thought (CoT) prompt. This CoT prompt is different from previous human created ones~\cite{kojima2022large} -- it does not instruct the LLM to think/write step by step, but instead assumes there already exists a solution with steps, that follows after the prompt.


\begin{table}[!ht]
    \centering
    \small
    \begin{tabular}{|p{0.94\linewidth}|}
        \hline
        \texttt{NQ:} ``\textit{Answer the question}'' $\rightarrow$ ``\textit{Compose a short, informative answer that directly answers the given question. The answer should be no longer than 15 words and should not contain any extraneous information. For example, if the question is "Who is the president of the United States?", the answer should be "Joe Biden". Do not write an essay or provide additional explanation.''}\\ 
        \hline
        \texttt{GSM8K:} ``SOLUTION"'' $\rightarrow$  ``\textit{Solve the problem by following the steps in the SOLUTION.}'' \\
        \hline
    \end{tabular}
    \caption{Prompt rewriting (initial prompt $\rightarrow$ rewritten prompt) for NQ and GSM8K produced by PRewrite-S. See Appendix~\ref{app:initial} and \ref{app:rewritten} for full results.}
    \label{tab:case}
\end{table}

Moreover, we find Prewrite always produces interpretable rewritten prompts, unlike RLPrompt, which often generates gibberish text. This is due to the LLM-based rewriter and KL-divergence penalty in PPO we have used (Section~\ref{sec:rl}).

\vspace*{-2pt}
\section{Related Work}
\label{sec:related_work}
\vspace*{-5pt}
We survey related work on automated prompt engineering for discrete prompts. Please refer \citet{liu2023pre} for a more comprehensive literature review.

Some earlier works optimize prompts via paraphrasing~\cite{Jiang2020,Yuan2021,Haviv2021}. In contrast, we adopt a powerful LLM to rewrite prompts, providing more capacity for prompt optimization. \citet{Shin2020,Wallace2021} propose gradient-based search approach which is challenging for larger API-access only models as it requires model gradient access.

Prior work have also explored RL based solutions. RLPrompt~\cite{Deng2022} optimizes prompts using RL, but often produces uninterpretable gibberish prompts. TEMPERA~\cite{Zhang2022} allows prompt editing at test time based on task input using RL, but it defines a small action space. By leveraging more capable LLMs, our method produces interpretable prompts and allows unconstrained exploration without a manually defined the action space. 

More recent works use blackbox LLMs such as PaLM 2 and GPT models, similar to ours. These include APE~\cite{Zhou2023}, Promptbreeder~\cite{Fernando2023} and OPRO~\cite{Yang2023}. These works use LLMs in different ways to propose prompt candidates and \emph{search} for the optimal one via validating performance on a given training dataset. We follow a similar idea but instead use RL for prompt optimization.

\section{Conclusions}
\label{sec:conclusion_future_work}
In this paper, we present PRewrite, a prompt rewriter trained with reinforcement learning (RL) for prompt optimization. We instantiate the rewriter with a LLM (PaLM 2-S) and finetune it using RL to optimize the end task performance. To further improve the performance, we develop a rewriting strategy that searches from the rewritten prompts generated by the trained rewriter. Our experiments testify the effectiveness of PRewrite and demonstrate its state-of-the-art performance.









\section{Limitations}
In this work, we only test with limited initial and meta prompts (see Table~\ref{tab:initial_prompts} and \ref{tab:meta_prompt} in Appendix) on four benchmark datasets. It would be interesting to experiment with more initial-meta prompt combinations to understand their implications, and on more datasets to test the generality of PRewrite. Moreover, we do not investigate the use of multiple meta/initial prompts to diversify exploration in prompt rewriting, which may further improve PRewrite. We leave these ideas for future work.

Due to resource constraints, we only experiment with PaLM 2 models. However, we believe that our conclusions should generalize to other LLMs as well.

\bibliographystyle{acl_natbib}
\bibliography{acl}
\clearpage
\appendix
\newpage
\appendix

\section{Problem Formulation Examples}
\label{app:problem_formulation_eg}
In Table~\ref{tab:def}, we show an example of the task input ($x$), output ($y$), prompt ($p$), and instruction ($t$) for a question answering task.

\begin{table}[!ht]
    \centering
    \small
    \begin{tabular}{|l|p{0.7\linewidth}|}
        \hline
        Input & \textit{Who is Harry Potter's father?}\\
        \hline
        Output & \textit{James Potter}\\
        \hline
        Prompt & \textit{Write a brief answer for the following question: Who is Harry Potter's father?}\\
        \hline
        Instruction & \textit{Write a brief answer for the following question}\\
        \hline
    \end{tabular}
    \caption{Example for a question answering task.}
    \label{tab:def}
\end{table}

\section{Meta Prompts}
\label{app:meta_prompt}
In Table~\ref{tab:meta_prompt}, we show the meta prompts used for prompting the rewriter LLM.
\label{app:meta_prompt}
\begin{table}[!ht]
    \centering
    \small
    \begin{tabular}{|p{0.93\linewidth}|}
        \hline
        \textit{Rewrite the following instruction via rephrasing and/or adding specific requirements. Use illustrative description if needed. Output the new instruction only.}\\
        \hline
        \textit{Rewrite the following instruction via rephrasing and/or adding specific requirements. Add instructions which would be helpful to solve the problem correctly. Output the new instruction only.}\\
        \hline
    \end{tabular}
    \caption{Meta prompts used for prompt rewriting, for experiments based on PaLM 2-S (upper) and PaLM 2-L (bottom) task model.}
    \label{tab:meta_prompt}
\end{table}

\section{Dataset Statistics}
\label{app:dataset_splits}
Data statistics for train/dev/test splits are given in Table \ref{table:dataset_splits}. GSM8K doesn't come with a dev (or validation) split, so we randomly reserve 10\% examples from the train split as the dev split.

\begin{table}[!ht]
\begin{center}
\small
\begin{tabular}{{lrrr}}
\toprule
Dataset & Train & Dev & Test \\
\midrule
AG News & 108,000 & 12,000 & 7,600\\
SST-2 & 60,614 & 67,35 & 871\\
NQ & 79,168 & 8,757 & 3,610\\
GSM8K & 6,725 & 748 & 1,319\\
\bottomrule
\end{tabular}
\end{center}
\caption{Train/Dev/Test splits for eval datasets.}
\label{table:dataset_splits}
\end{table}

\section{Results based on Different Rewards}
\label{app:results_rewards}
We test different rewards for all datasets and report the results based on PRewrite-I and PaLM 2-S task model in Table~\ref{tab:reward}. Perplexity uses perplexity of the ground truth labels as the reward. F1 use word-level F1 measure as the reward. Perplexity+F1 sums perplexity and F1 as the reward.    

\begin{table}[!ht]
    \centering
    \small
    \begin{tabular}{lccc}
        \toprule
        Reward  &  AG News  &  SST-2  &  NQ \\
        \midrule
        EM/Accuracy  &  84.5  &  96.5  &  29.3 \\
        F1  &  84.5  &  96.6  &  30.6 \\
        Perplexity  &  60.1  &  95.8  &  12.7 \\
        Perplexity+F1  &  84.2  &  96.5  &  32.3 \\
        \bottomrule
    \end{tabular}
    \caption{PRewrite-I experiment results based on different rewards and PaLM 2-S task model.}
    \label{tab:reward}
\end{table}

First, we find that using the final task metric, accuracy or EM, performs well in general. In other words, RL is able to directly optimize for these task metrics.
Second, we find that F1 is in general be more stable than accuracy/EM. This is because F1 can provide more fine-grained feedback for RL -- accuracy/EM is either 0 or 1 for a datapoint while F1 provides a fractional score. 
Third, we find perplexity can sometimes be harmful (see NQ and AG News results in the table), as it is not directly linked to the final task metrics. However, combining perplexity with F1 gives the best performance on NQ.

\section{Initial Prompts}
\label{app:initial}
Table~\ref{tab:initial_prompts} lists all initial prompts we used for prompt rewriting in our experiments.

\begin{table*}[!ht]
    \centering
    \small
    \begin{tabular}{lcp{0.6\linewidth}}
        \toprule
Dataset &  Source  & Initial prompt  \\
\midrule
AG News  & -   &  \textit{Given a news article, categorize it into one of the following categories: 1. World 2. Sports 3. Business 4. Sci/Tech.} \\
\hline
SST-2 & \citet{Zhang2022} & \textit{In this task, you are given sentences from movie reviews. The task is to classify a sentence as “positive” if the sentiment of the sentence is positive or as “negative" if the sentiment of the sentence is negative.}\\
\hline
NQ & - & \textit{Answer the question}\\
\hline
GSMK & \citet{Fernando2023} & \textit{SOLUTION"}\\
\hline
    \end{tabular}
    \caption{Initial prompts used for prompt rewriting in our experiments. When the source is absent, the prompt is manually crafted by us.}
    \label{tab:initial_prompts}
\end{table*}

\begin{table*}[!ht]
    \centering
    \small
    \begin{tabular}{lcp{0.6\linewidth}}
        \toprule
        Dataset &  Strategy & Rewritten prompt  \\
        \midrule
        AG News  & PRewrite-I & \textit{Classify a news article into one of the following categories: World, Sports, Business, Sci/Tech.} \\
        \hline
        AG News & PRewrite-S & \textit{Classify a given news article into one of the following categories: World, Sports, Business, or Sci/Tech.}\\
        \hline
        SST-2  & PRewrite-I & \textit{sentiment classification} \\
        \hline
        SST-2 & PRewrite-S & \textit{sentiment classification from text}\\
        \hline
        NQ & PRewrite-I & \textit{Compose a short, informative answer to the given question. The answer should be no longer than 15 words and should be written in a clear, concise manner. For example, if the question is "Who is the president of the United States?", the answer should be "Joe Biden". Do not write an essay or provide additional explanation.}\\
        \hline
        NQ & PRewrite-S & \textit{Compose a short, informative answer that directly answers the given question. The answer should be no longer than 15 words and should not contain any extraneous information. For example, if the question is "Who is the president of the United States?", the answer should be "Joe Biden". Do not write an essay or provide additional explanation.}\\
        \hline
        GSM8K & PRewrite-I & \textit{Provide a detailed solution to the problem.} \\
        \hline
        GSM8K & PRewrite-S & \textit{Provide a solution to the problem in a clear and concise manner.}\\
        \bottomrule
    \end{tabular}
    \caption{Rewritten prompts produced by PRewrite based on PaLM 2-S task model.}
    \label{tab:rewritten_prompts_s}
\end{table*}

\begin{table*}[!ht]
    \centering
    \small
    \begin{tabular}{lcp{0.6\linewidth}}
        \toprule
        Dataset &  Strategy & Rewritten prompt  \\
        \midrule
        GSM8K & PRewrite-S & \textit{Solve the problem by following the steps in the SOLUTION.}\\
        \bottomrule
    \end{tabular}
    \caption{Rewritten prompts produced by PRewrite based on PaLM 2-L task model.}
    \label{tab:rewritten_prompts_l}
\end{table*}

\begin{table*}[!ht]
    \centering
    \small
    \begin{tabular}{lp{0.6\linewidth}}
        \toprule
    Dataset & Prompt template  \\
    \midrule
    AG News  & \texttt{"\{$t$\}\textbackslash nArticle: \{title\} \{description\}"} \\
    \hline
    SST-2  & \texttt{"\{$t$\}\textbackslash nText: \{text\}"} \\
    \hline
    NQ  & \texttt{"\{$t$\}\textbackslash nQuestion: \{question\}"} \\
    \hline
    GSM8K  & \texttt{"\{$t$\}\textbackslash nQuestion: \{question\}"} \\
    \bottomrule
    \end{tabular}
    \caption{Prompt templates used for each datasets. $t$ is the initial/rewritten task instruction.}
    \label{tab:prompte_templates}
\end{table*}

\section{Rewritten Prompts}
\label{app:rewritten}
Table~\ref{tab:rewritten_prompts_s} lists the produced rewritten prompts for experiments using PaLM 2-S as the task LLM. This includes rewritten prompts produced using both rewriting strategies (Section~\ref{sec:search}) for all datasets.

Table~\ref{tab:rewritten_prompts_l} lists the rewritten prompts produced by PRewrite-S for GSM8K experiments using PaLM 2-L as the task LLM.

\section{Prompt Templates}
\label{app:prompt_templates}
Table~\ref{tab:prompte_templates} lists the prompt template we used for each dataset.

\end{document}